\documentclass{article} 
\usepackage[a4paper, total={6in, 8in}]{geometry}
\usepackage{hyperref}

\usepackage{times}  
\usepackage{helvet}  
\usepackage{courier}  
\usepackage[square,sort,comma,numbers]{natbib}  
\usepackage{caption} 
\usepackage{amsmath}
\DeclareCaptionStyle{ruled}{labelfont=normalfont,labelsep=colon,strut=off} 
\usepackage{algorithm}
\usepackage{algorithmic}

\usepackage{newfloat}
\usepackage{listings}
\floatstyle{ruled}
\newfloat{listing}{tb}{lst}{}
\floatname{listing}{Listing}
\usepackage{natbib}

\usepackage{mathtools}
\usepackage{amssymb,amsmath,amsthm}
\usepackage{soul}
\usepackage{url}            
\usepackage{xcolor}
\usepackage{xr-hyper,refcount}

\usepackage[utf8]{inputenc} 
\usepackage[T1]{fontenc}    
\usepackage{url}            
\usepackage{booktabs}       
\usepackage{amsfonts}       
\usepackage{nicefrac}       
\usepackage{microtype}      
\usepackage{amsmath}
\usepackage{graphicx}
\usepackage{subfigure}
\usepackage{algorithm}
\usepackage{algorithmic}
\usepackage{caption}

\title{Improving Classifier Robustness through Active Generation of Pairwise Counterfactuals}

\author{Ananth Balashankar\thanks{Corresponding author: ananthbshankar@google.com}, Xuezhi Wang, Yao Qin, Ben Packer, Nithum Thain,\\ Jilin Chen, Ed H. Chi, Alex Beutel\\
Google Research\\USA}

\begin{document}

\maketitle

\begin{abstract}
Counterfactual Data Augmentation (CDA) is a commonly used technique for improving robustness in natural language classifiers. However, one fundamental challenge is how to discover meaningful counterfactuals and efficiently label them, with minimal human labeling cost.
Most existing methods either completely rely on human-annotated labels, an expensive process which limits the scale of counterfactual data, or implicitly assume label invariance, which may mislead the model with incorrect labels. 
In this paper, we present a novel framework that utilizes counterfactual generative models to generate a large number of diverse counterfactuals by actively sampling from regions of uncertainty, and then automatically label them with a learned pairwise classifier. 
Our key insight is that we can more correctly label the generated counterfactuals by training a pairwise classifier that
interpolates the relationship between the original example and the counterfactual.
We demonstrate that with a small amount  of human-annotated counterfactual data (10\%), we can generate a counterfactual augmentation dataset with learned labels, that provides an 18-20\% improvement in robustness and a 14-21\% reduction in errors on 6 out-of-domain datasets, comparable to that of a fully human-annotated counterfactual dataset for both sentiment classification and question paraphrase tasks. 
\end{abstract}

\section{Introduction}
Counterfactual data augmentation (CDA) has been used to make models robust to distribution shift and mitigate biases towards spuriously correlated attributes. Often, counterfactuals are generated as labeled examples through pre-specified templates \cite{dixon, hall-maudslay-etal-2019-name} or crowd-sourcing \cite{Kaushik2020Learning}. While natural text templates codify a specific number of assumptions of how counterfactual sentences and labels might vary, crowd-sourcing that can cover various types of counterfactuals, can be expensive. 
On the other hand, many existing methods \cite{fairgan,natural_gan, jia-etal-2019-certified, alzantot-etal-2018-generating} simply rely on 
a label-invariance assumption: the label of the generated counterfactual example is the same as the corresponding original example. However, this simple label-invariance assumption does not always hold \cite{ribeiro-etal-2020-beyond, pmlr-v119-tramer20a, DBLP:journals/corr/abs-2009-10195} and thus greatly increases the risk of using incorrect labels for counterfactuals during training. 
For example, for many NLP tasks a small perturbation can easily change the ground-truth label \cite{Kaushik2020Learning,DBLP:journals/corr/abs-2004-02709}, e.g., changing the input from \textit{This movie is great} to \textit{This movie is supposed to be great} for sentiment classification, or changing the hypothesis from \textit{The lady has three children} to \textit{The lady has many children} for natural language inference.
Therefore, in this work, we mainly focus on addressing this challenging research problem:

\emph{``How can we automatically explore diverse counterfactual examples and learn their labels, given a counterfactual text generator?''} 

Beyond costly human annotation or simplifying assumptions of label invariance, researchers have explored to use a classifier $f$ that has learnt to predict the label on the original dataset $(X, Y)$. Such a classifier has been used to directly label generated examples (our ``trust'' baseline;  \cite{Kaushik2020Learning}) or to weight generated examples based on the model uncertainty (our weighted-trust baseline; \cite{ovadia2019trust}).  However, we see that using such simplistic labeling assumptions for counterfactual data augmentation have limited benefits for improving robustness (defined as the accuracy over a counterfactual test set of interest).

\begin{figure}[h]
\centering
\vspace{-0.1in}
\includegraphics[width=0.48\textwidth]{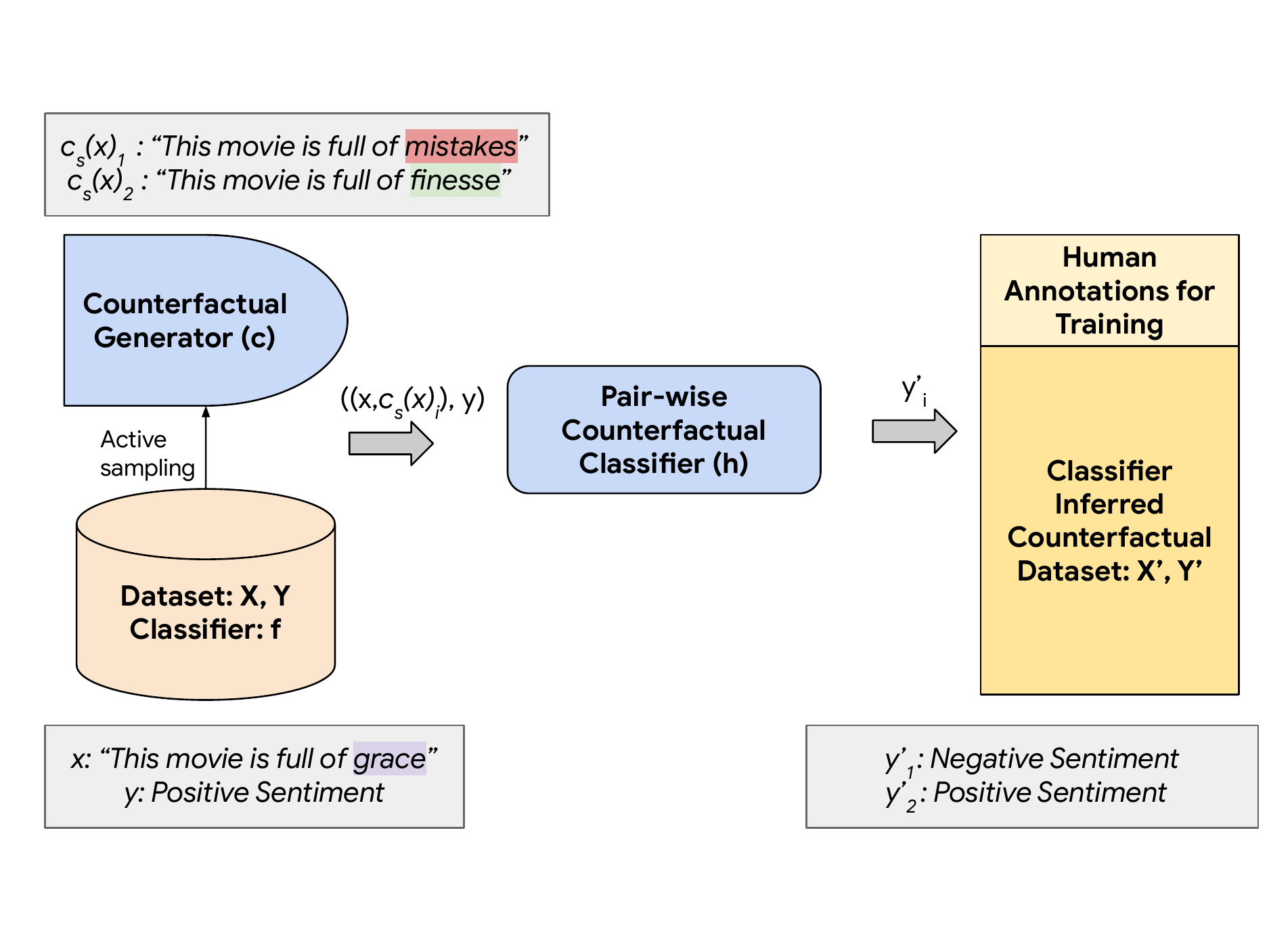}
\vspace{-0.2in}
\caption{\textbf{Overview of proposed approach:} 
We propose to generate diverse counterfactuals through active sampling, and label them using a pairwise counterfactual classifier at scale. We use the labeled counterfactuals as data augmentation over the original classifier (not shown) to significantly improve robustness.}
\label{schema}
\end{figure}

In this paper we propose an alternative approach to this problem: we leverage the sample efficiency of generative models and exploration capabilities of active learning \cite{cohn1996active} to 1) first generate a large number of {diverse} counterfactuals, and 2) then train an {auxiliary classifier} to automatically annotate the generated counterfactual data based on the {difference} between the original and counterfactual labels. Specifically, we propose to generate counterfactual examples that lie in the region of uncertainty of the classifier $f$, and learn a pairwise classifier $h$ to predict the counterfactual label $y'$. The pipeline of our method is shown in Figure \ref{schema}. 

In particular, we utilize a very small set of human-annotated counterfactual examples to train the pairwise counterfactual classifier $h$, which takes in the {pair} of original and counterfactual sentences $(x, c_s(x))$ and the original label $y$ as input. Then in the inference stage, the pairwise counterfactual classifier $h$ is used to predict the labels to produce a large counterfactual augmentation dataset used to fine-tune $f$ to improve robustness. 


By using active sampling over
counterfactual generators and auxiliary pairwise counterfactual classifiers, we show that we greatly reduce 
the number of counterfactual examples for which we need human annotation, while providing similar gains in robustness comparable to a fully human annotated counterfactual dataset. We attribute this to two core components of our method. 
First, the active-learning based sampling method helps diversify the types of generated counterfactuals and enables them to capture different robustness issues not previously captured by our classifier model. Second, the proposed auxiliary pair-wise classifier can automatically annotate the generated counterfactuals with more accurate labels and help efficiently scale up the size of the counterfactual augmentation dataset.
Our core contributions in this work include:
\begin{itemize}
    \item We propose an active-learning based sampling method to generate diverse counterfactuals and effectively improve robustness of classifiers for the sentiment classification task on Stanford Sentiment Treebank (SST-2) dataset, and the question paraphrase task on Quora Question Pair (QQP) dataset.
    \item We propose a novel pairwise counterfactual classifier to automatically label counterfactually generated examples based on a small set of annotated counterfactuals,  improving sample efficiency of counterfactual data augmentation.
    \item The generated augmented dataset, which uses just 10\% of human-annotated labels, produces an improvement in counterfactual robustness of 18-20\%, comparable to a fully human annotated dataset, and a reduction in errors by 14-21\% on out-of-domain datasets that were not used during training: IMDB, Amazon, SemEval (Twitter), and Yelp reviews.
\end{itemize}

\section{Related Work}
Our work is built on advances from various domains as outlined below:

\textbf{Adversarial Text Generation}
Training against adversarial examples which perturb inputs in the vicinity of the existing training data by making geometric assumptions \cite{DBLP:journals/corr/abs-2009-10195, zeng2021certified} on a lower dimensionality of the data to improve robustness has been extensively studied recently. Natural examples which are syntactically and semantically similar to the original sentence, but produce different model predictions have been produced \cite{alzantot-etal-2018-generating}. Similarly, defenses against adversarial attacks on self-attentive models have shown improvement in robustness to label invariant examples \cite{hsieh-etal-2019-robustness}. In FairGAN \cite{fairgan}, they showed it is possible for a discriminator to achieve statistical parity on the real dataset, while performing the auxiliary task of detecting real and generated examples. Such controlled adversarial generative approaches \cite{wang-etal-2020-cat} have demonstrated the effectiveness of automating data augmentation in text-based tasks. Generative models which optimize for fluency have passed human annotation checks where the model generated text is almost indistinguishable from human generated ones \cite{DBLP:journals/corr/abs-2012-04698, DBLP:journals/corr/abs-2012-13985}. We build on this body of work and utilize a generative model \cite{polyjuice:acl21} that captures template-based counterfactuals to improve robustness. Generic adversarial notions of robustness however applicable, fail to incorporate specific counterfactuals directly in their training and orthogonal to our scope of study. Through carefully disentangling specific attributes and the rest of the latent variables in text, we generate counterfactuals across all possibilities, and utilize human-annotated templates to label a small fraction of the generated examples to train a pairwise counterfactual classifier.

\textbf{Semi-Supervised and Self-Supervised Learning} Labeling functions which provide crude estimates of the label have been used in semi-supervised methods \cite{scudder1965probability, chapelle2009semi, DBLP:journals/corr/abs-1711-10160}, and are further used to learn a generative model to generalize over them. Further, utilizing unlabeled data \cite{carmon2019unlabeled} to improve adversarial robustness leverages geometric smoothing-based techniques to bridge the sample complexity gap between accuracy and robustness \cite{DBLP:journals/corr/abs-2003-02460}. Thus, semi-supervised learning approaches aim to generate examples where the discriminator is least confident about \cite{ovadia2019trust}. Language models with very large number of parameters have also shown to be few-shot learners with minimal supervision \cite{brown2020language}. Similarly, reinforcement learning based approaches with minimal labels have been proposed to combine the objectives of accuracy and counterfactual robustness \cite{pitis2020counterfactual}. Generalization against counterfactual examples by making models not to rely on salient features (easy examples) have been extensively studied by modeling biases in corpora \cite{clark-etal-2019-dont, clark-etal-2020-learning, karimi-mahabadi-etal-2020-end, DBLP:journals/corr/abs-2010-12510, utama-etal-2020-towards, ghaddar-etal-2021-end}. While the goal in these works have been building ensembles or end-to-end bias mitigation models, our goal is to minimize the number of human labels required to achieve an equivalent improvement in robustness. In this spirit of efficiently capturing the patterns already prevalent in the original dataset, and learning only the new ones introduced in the counterfactual templates, we learn the pairwise counterfactual classifier on a small number of samples, and use it to capture the label variations in the remaining counterfactual dataset.

\textbf{Counterfactual Applications}
The counterfactual datasets we use throughout this paper were intended to highlight the shortcomings of existing models at the time. Improving robustness through training on the augmented data has been extensively explored \cite{10.1145/3306618.3317950, wu2018conditional}. Learning how counterfactuals differ have been explored by comparing against gradient supervision \cite{teney2020learning} and the generalizability between original and counterfactuals \cite{Kaushik2020Learning}. The generated counterfactuals have also been used for explanations \cite{verma2020counterfactual}, highlighting biases \cite{dixon} and debiasing through statistical methods \cite{lu2019gender}. This rich set of contrast sets \cite{DBLP:journals/corr/abs-2004-02709}, checklists \cite{ribeiro-etal-2020-beyond}, paraphrases \cite{zhang2019paws, wieting-gimpel-2018-paranmt}, adversarial schemes \cite{sakaguchi2019winogrande} and lexical diagnostic datasets \cite{mccoy-etal-2019-right} form the foundation of our method, which re-purposes them to build a counterfactual generative model and improve counterfactual robustness.

\textbf{Generative Learning}
Generative adversarial active learning has been proposed with pool-based and synthesizing-based sampling strategies \cite{zhu2017generative}. While the pool-based strategy selects from an existing sample of generated examples, the synthesizing-based sampling re-samples from the generator based on information theoretic measures like mutual information \cite{houlsby2011bayesian, tran2019bayesian} or model uncertainty \cite{gal2017deep} or informativeness \cite{mackay1992information, cohn1996active} over the initial sample. Further, recent work has highlighted that using large language models for generation and annotation of that generated data can be very useful \cite{gilardi2023chatgpt, liu-etal-2022-wanli}. We build on this work, and use the synthesizing-based Bayesian Generative Active Learning approach \cite{tran2019bayesian}, where we condition on a counterfactual with low uncertainty in the model, to actively generate more counterfactual examples. We then iteratively sample the examples with the highest classifier uncertainty and annotate manually. The human annotations are then used to train the pairwise classifier $h$, which is then used to scale the annotation process for all other generated counterfactuals.

\section{Methodology}

\subsection{Problem Framing}
Let $x$, $y$ be the input sentence and its associated label in the original dataset, respectively. 
We assume $y\in \{0, 1\}$ throughout the paper (i.e., we focus on binary classification tasks), but our framework can be extended to multi-class tasks as well.

Our core challenge is what is the true label $y'$ for a generated  counterfactual $x'$?  Although we can further obtain human annotations, this can quickly become time consuming and budget intensive to do at scale. If we make the simplified assumption  of label invariance throughout the counterfactual inputs $x’$ generated, which is a common assumption in adversarial literature \cite{goodfellow15, jia-etal-2019-certified, alzantot-etal-2018-generating}, we could end up with an incorrect counterfactual dataset which might hurt robustness and accuracy.  Our goal is thus, to \textit{generate a counterfactual augmentation dataset that produces a comparable improvement in accuracy and robustness as that of human-annotated counterfactuals with minimal supervision}. 

We frame this problem as how to learn when the labels flip, i.e., identifying when the label of the counterfactual is different from the label of the original sentence:
$P(y \neq y’) = \delta$, $(0 < \delta < 1)$, in the counterfactual distribution $x’ \in X’$. 
Given a generation model $c$, we denote $c_s(x)$ as the generated counterfactual over $x$ by changing an attribute $s$ in $x$. We also assume that a classifier $f: X \rightarrow Y$ has been learnt on the original dataset $(X, Y)$ by optimizing for accuracy $A$.
\begin{align}
    A =& E_{(x,y) \in (X,Y)} \mathbb{I} (f(x) = y)
\end{align}
In our paper, the objective is to use the counterfactual data to train a model $f'$ that improves robustness, i.e., to make sure the models we trained generalize to unseen scenarios.  We measure this by the counterfactual accuracy $\Tilde{A}$ of $f$ on multiple held-out counterfactual datasets $(X', Y')$ split based on domains (OOD), patterns (e.g. negation, insertion):
\begin{align}
    \Tilde{A} =& E_{(x',y') \in (X',Y')} \mathbb{I} (f'(x') = y') \label{robust}
\end{align}

In the remainder of this section, we first explain how the counterfactuals are generated using active learning. Then, we explain how we account for the possibility that the label of the generated counterfactual $c_s(x)$ might have flipped, using a pairwise classifier. Finally, we explain how when both these components are combined, we can further improve robustness.


\subsection{Active Counterfactual Generation}
To achieve the goal of improving counterfactual accuracy on held-out counterfactual datasets (Eqn \ref{robust}), we use a controlled generative model to generate additional training counterfactual data $c_s(x) \in X_t'$ (here the subscript $t$ denotes the training set) that modifies original input $x \in X$ based on the attribute $s$. In natural language tasks, the attribute $s$ cannot be directly inferred from the sentence $x$ and hence we rely on templates to define the types of counterfactual (e.g., negation, insertion, deletion) as commonly used in \cite{ribeiro-etal-2020-beyond, polyjuice:acl21} to infer the attribute $s$. Let $y \in Y,y' \in Y_t'$ be the label for the original and counterfactual sentences in our counterfactual training dataset. The training objective of robustness is to minimize the error $\Tilde{\mathcal{E}}_t$ of the model $f$ aggregated by attribute $s$ on the training counterfactuals $(X_t',Y_t')$, where $CE$ refers to the cross-entropy loss, as follows:
\begin{align}
\hspace{-0.1in}    \Tilde{\mathcal{E}}_t(s) &= E_{x \in X, (c_s(x),y') \in (X_t',Y_t')} CE(f(c_s(x)), y')\\
    \Tilde{\mathcal{E}}_t &= E_{s \in S} \Tilde{\mathcal{E}}_t(s)
\end{align}

The counterfactual generator that optimizes the above cross-entropy loss then generates several counterfactuals $c_s(x)$ by relying on instructions provided in controlled generation methods \cite{polyjuice:acl21} such as ``negation'', ``restructure''. However, these counterfactuals are not necessarily diverse, and fails to incorporate the classifier's uncertainty to get the most informative set of generated counterfactuals.
To improve generalization across a diverse set of counterfactual types, we fine-tune the generator to actively sample counterfactuals the most informative set from the unlabeled dataset $x* \in X_t'$ (BALD, \cite{houlsby2011bayesian}) that synthesizes examples by maximizing the acquisition function given by the Monte Carlo (MC) dropout approximation method \cite{gal2017deep} using the class-wise probability scores of the pairwise classifier $h$, where $H[y|x,.]$ represents the Shannon entropy of the corresponding conditional probability:
\begin{align}
\label{genactive}
    x^* = arg max_{x' \in X_t'} & [H[y'| x', X, Y] \\\nonumber
     - & \mathbb{E}_{x \in X_t} H[y'| x', x, f(x)]] 
\end{align}

Since $y'$ is not readily available for counterfactual generated sentences $c_s(x)$ in our training dataset and gathering them for all examples can be expensive, our goal is to minimize the number of human-annotations of counterfactuals $y'$ in the training dataset $Y_t'$, while achieving comparable improvement in robustness (Eqn \ref{robust}). Hence, the training sentence and label set $(X_t', Y_t')$ can be decomposed into two sets, one whose labels are human-annotated: $(X_a', Y_a')$ and the other with model generated labels: $(X_g', Y_g')$, such that $X_t' =  X_a' \cup X_g', Y_t' = Y_a' \cup Y_g'$. Our goal is to automatically discover informative counterfactuals $X'_g$ and learn their labels with access to a limited human-annotated counterfactual data ($X'_a, Y'_a$), where $|Y'_a|\ll |Y'_g|$, while achieving counterfactual robustness $\Tilde{A}$ (Eqn \ref{robust}) comparable to the scenario when all the training labels are human-annotated.

\subsection{Pairwise-Counterfactual (PC)}
In order to generate labels for the counterfactuals, we construct a novel \textit{auxiliary pairwise classifier} $h$, which at inference time, takes in as input both the original dataset $(x,y) \in (X,Y)$, and a corresponding counterfactual $c_s(x) \in X_g'$, to output $y'\in Y_g'$. This classifier $h$ is trained on \emph{pairs} of input sentences $x, c_s(x)$ and the original label $y$ to predict the human-annotated label $y' \in Y_a'$.

Specifically, the classifier $h$ takes in the original input sentence $x$ and its associated label $y$, as well as its corresponding counterfactual example $c_s(x)$. The output of the classifier $h(x, c_s(x), y)$ is the predicted label of the counterfactual example $c_s(x)$. In the training stage, the classifier $h$ is optimized on the counterfactual examples with human-annotated labels $(c_s(x), y') \in (X_a', Y_a')$ via minimizing the loss function:
\begin{equation}
 \ell_h = E_{\substack{(x,y) \in (X, Y) \\ (c_s(x), y') \in (X_a', Y_a')}} CE (h(x, c_s(x), y), y')  
\end{equation}

With the well-trained classifier $h$, we can generate the labels for any counterfactual example $c_s(x) \in X'_g$ (the counterfactual set without human annotation) according to:
\begin{equation}
y' =  h(x, c_s(x), y): (x,y) \in (X,Y), c_s(x) \in X_g' 
\end{equation}

\subsection{Classifier-Aware Pairwise-Counterfactual (CAPC)}
Additionally, since we know that $f$ is already optimized to predict the label accurately on the original dataset, the auxiliary classifier $h$ could potentially leverage $f$ in its pairwise prediction through transfer learning. Specifically, if we decompose the counterfactual distribution $(X', Y')$ as a mixture of samples from the original distribution $(X, Y)$ and those that are independent of the original distribution, we would benefit by training $h$ to identify samples from the latter distribution. 
In addition, assuming the correspondence between $f(x)$ and $f(c_s(x))$ is easier to learn (e.g., with a lower model complexity), we could also benefit from learning a classifier-aware function to better capture this correspondence.
Thus, we propose to augment the predictions of the original classifier $f(x),f(c_s(x))$ as input to $h$ as follows:
\begin{align}
\label{capc_eq}
    y' \in Y_g' = h(x, c_s(x), y, f(x), f(c_s(x))): \\\nonumber(x,y) \in (X,Y), c_s(x) \in X_g'
\end{align}
Any uncertainty that $f$ has on the counterfactual samples $P(f(c_s(x)) \neq y')$ can be mitigated by the auxiliary classifier $h$ by identifying patterns in $c_s(x)$ when $f$ predicts incorrectly. As a simple example, without any human annotation, the original model $f$ might make incorrect assumptions on $c_s(x)$ that lead to incorrect predictions $f(c_s(x))\neq y'$, e.g., a sentiment analysis model might give ``positive'' sentiment predictions due to the presence of qualifiers like ``terrific'', ``amazing'' (\emph{this movie was amazing})  even when the counterfactual input $c_s(x)$ alters aspects of a sentence that changes the label (\emph{this movie was supposed to be amazing}). But, this can be corrected using Eqn \ref{capc_eq} after $h$ has observed some data over the correct correlation between $x, c_s(x), y, f(x), f(c_s(x))$ and $y'$, especially if there exists a lower-complexity function mapping between them - for instance, adding the phrase ``supposed to be" may alter the label of a review.

\section{Evaluation}
We evaluate on two NLP tasks, sentiment classification and question paraphrase, using two datasets namely the Stanford Sentiment Treebank (SST-2) \cite{socher-etal-2013-recursive} and the Quora Question Pair (QQP) \cite{qqp_data, wang-etal-2018-glue}. 
When the CAPC classifier is used in conjunction with generated examples through active learning, we correspondingly prefix the model name as \textbf{p-CAPC} (pool-based sampling with no retraining as per Eqn \ref{genactive}) or \textbf{s-CAPC} (examples synthesized with re-generation of counterfactuals optimizing Eqn \ref{genactive}).

\subsection{Counterfactual Generator: Polyjuice}
We use a general purpose counterfactual text generator called Polyjuice \cite{polyjuice:acl21}, which extends CheckList \cite{ribeiro-etal-2020-beyond}, that has shown promise by improving diversity, fluency and grammatical correctness as evaluated by user studies. It covers a wide variety of commonly used counterfactual types including patterns of negation \cite{Kaushik2020Learning}, adding or changing quantifiers \cite{DBLP:journals/corr/abs-2004-02709}, shuffle key phrases \cite{zhang2019paws}, word or phrase swaps which do not alter POS tags \cite{sakaguchi2019winogrande} or parse trees \cite{wieting-gimpel-2018-paranmt}, along with insertions or deletion of constraints that do not alter the parse tree \cite{mccoy-etal-2019-right}.  Specifically, we use 8 types of counterfactuals - negation, quantifier, lexical, resemantic, insert, delete, restructure, shuffle; in Polyjuice to generate the augmented dataset.
Other text generative models like \cite{natural_gan, Kaushik2020Learning, jia-etal-2019-certified} that improve adversarial robustness or like \cite{keskar, dathathri} that allow controlled generation could be used as well. 

\subsection{Experiment Setup}
We test our methods on two popular text datasets. We briefly describe the two datasets below, and discuss the different evaluations of counterfactual robustness we perform over them.

\textbf{Stanford Sentiment Treebank (SST-2):}
The sentiment analysis task in SST-2 \cite{socher-etal-2013-recursive} assigns a binary sentiment (negative/positive) to a sentence mined from RottenTomatoes movie reviews. The corresponding counterfactuals are generated using the Polyjuice generator \cite{polyjuice:acl21}. The original dataset contained 4,000 samples, while the counterfactual dataset had 2,000 samples with human labels against which we evaluate.
We show a sample of the dataset in the following:
\\
\vspace{0.1in}
\fbox{%
{\scalebox{0.90}{
\parbox{\columnwidth}{ 
{ \textbf{Positive}: A dog is embraced by the dog
} 
\\
{ \textbf{Negative}:}  A dog is not embraced by the dog
}
}
}
}
\vspace{0.1in}

\noindent
\textbf{Quora Question Pair:} 
In the QQP dataset \cite{qqp_data, wang-etal-2018-glue}, given a pair of questions, the task is to predict if they are semantically equivalent, hence marked as duplicate. Here, again the second question is modified by Polyjuice \cite{polyjuice:acl21} as per the templates used for the SST-2 dataset including negation, insertion, deletion, rephrasing, etc, out of which 1,911 samples were human annotated for evaluation. The original dataset had 20,000 samples.\\
\vspace{0.1in}
\fbox{%
{\scalebox{0.90}{
\parbox{\columnwidth}{ 
{ \textbf{Duplicate}: How can I help a friend experiencing serious depression?; How can I help a friend who is in depression?
} 
\\
{ \textbf{Non-duplicate}:}  How can I help a friend experiencing serious depression?; How can I play with a friend who is in depression?
}
}
}
}
\\
\textbf{Evaluation:}
In both datasets, we have a small number of counterfactual human annotations available (SST-2: 2,000; QQP: 1,911) \cite{polyjuice:acl21}. We divide these examples into two sets, one for training and annotating using $h$, and another held-out test dataset used to compute counterfactual robustness of $f$. The former dataset is used for fine-tuning $f$ for counterfactual robustness,  while the latter is used only as a held-out test set. In the SST-2 dataset, this means we split out 1,000 samples for training/annotation and 1,000 as the test set, while in the QQP dataset, we use 1,000 samples for training/annotation and the remaining 911 samples for testing counterfactual robustness. However, our aim is to use a minimal subset of the 1,000 samples available for training the base classifier directly. Instead, we use a smaller training dataset (100) to train our pairwise classifier which in-turn can then \emph{artifically} annotate the remaining (say 900) samples. The combination of these (sum to 1000) will then be used to train the base classifier. Thus, in all our experiments, the number of counterfactual samples available to the base classifier to train on remains the same, although at different levels of human labeling costs. 

The classifier $f$ is first trained on the original classifier and then fine-tuned on the counterfactual dataset. We also perform 10 random initializations of the model $f$ and $h$ and a 10-fold cross-validation split on the training/annotation data, thus report the mean and standard error bounds $\sigma/\sqrt{n}$ over $n=1000$ runs for each model-based annotation and training for counterfactual robustness. We used the standard hyperparameters provided in RoBERTa
for training $f$ on $(X, Y)$ and the hyperparameters for fine-tuning $f$ on $(X_t', Y_t')$ include learning rate of $5e^{-5}$, batch size of $16$ and a sequence length of 120 for 20 epochs. The pairwise counterfactual classifier's hyperparameters were chosen after a grid search to have a learning rate of $5e^{-4}$, batch size of $32$ for 50 epochs, sequence length of $240$  including the original label and classifier predictions with special marker characters. While the base classifier $f$ is trained on contextual embeddings of the sentence(s), $h$ is trained by further augmenting the original and counterfactual sentence embeddings as input to RoBERTa followed by the base classifier’s predictions separated by special delimiters [DEL]. A similar 10-fold cross-validation split is used to finetune the parameters of the classifier $h$.

\textbf{Out-of-Distribution (OOD)}: To test the methodology on out-of-domain datasets, we test on sentiment analysis tasks in 6 reviews datasets - IMDB movie (3 including contrast sets) reviews, Amazon, SemEval, and Yelp reviews \cite{kaushik2021learning}. The IMDB reviews (1,700) were collected by \cite{Kaushik2020Learning} through careful human elicitation to produce label varying counterfactuals of existing IMDB reviews. In the Yelp reviews \cite{DBLP:journals/corr/Asghar16}, the task is to predict the ratings of 115,907 reviews on a scale of 1-5, and in the Amazon reviews \cite{ni-etal-2019-justifying}, we evaluate on the 57,947 reviews in the clothing product category. Each of these datasets was not used for training either the base classifier or the pairwise classifier, and the training relies solely on the SST-2 dataset. So, we can measure the generalizability of the pairwise classifier based data augmentation methodology.

\subsection{Baselines}
We now briefly describe five different baselines used to generate the labels of counterfactual augmented data ($Y_g'$), given access to a small number of annotated labels $Y_a'$. 
\begin{itemize}
    \item
    \textbf{No-cda}: $f$ without any counterfactual data used for robustness.
    \item \textbf{Label-invariant (invariant)} : the labels of the counterfactual examples are assumed to be the same as the original sentence: $y' =y$ (except for the counterfactuals generated for the negation type, where it is the opposite).
    \item  \textbf{Trust}: we trust the classifier $f$ to annotate the counterfactual labels $y' = f(c_s(x))$ - a form of semi-supervision based on the existing base classifier.
    \item \textbf{Weighted-trust} (\textbf{w-trust}): the label of the counterfactual example is computed via the maximum score weighted by the confidence score of the classifier $f$ on the pair for a label $l:$ $p_l(x)$ such that  $y' = \arg\max_l p_l(x) \cdot p_l(c_s(x))$. 
    \item \textbf{Random}: In order to understand the importance of the counterfactual sentences used in the pairwise classifier, we also evaluate against a classifier which takes two randomly paired sentences from the original dataset as input and predicts the second label given the label of one sentence.
    \item \textbf{Training}: we only use those counterfactual examples with human-annotated labels ($X'_a, Y'_a)$ and drop all other counterfactual examples.
\end{itemize}


For all these baselines as well as our proposed methods, we use the RoBERTa \cite{roberta} fine-tuned model as the choice of classifier $f$, and a corresponding pairwise fine-tuning task using RoBERTa \footnote{\label{note1}huggingface.co/roberta-large-mnli,
textattack/roberta-base-SST-2,
ji-xin/roberta\_base-QQP-two\_stage} for the auxiliary pairwise counterfactual classifier $h$.

 \section{Results}
\subsection{Improving Counterfactual Robustness}
\begin{figure}[htbp]
\centering
\includegraphics[width=\columnwidth]{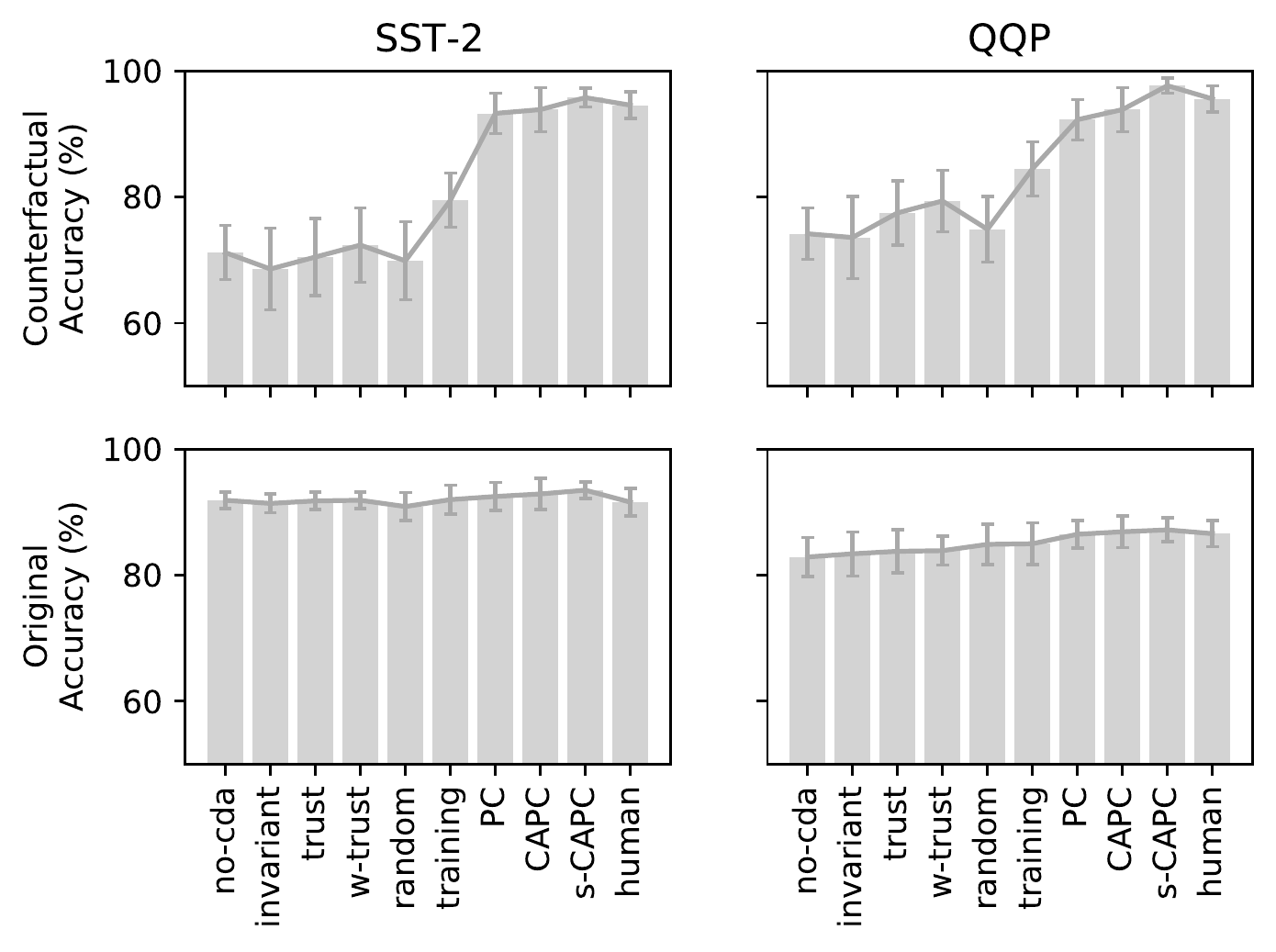}
\caption{\textbf{(a) Robustness:} (first row) Training on 10\% of human-annotated counterfactuals, and annotating the rest using the auxiliary classifier, we achieve a comparable improvement in robustness (lower error rate) for both Stanford Sentiment and Quora Question Pair datasets; \textbf{(b) Accuracy: } This improvement in robustness does not sacrifice the accuracy on the original held-out dataset.}
\label{robust1}
\end{figure}

To demonstrate the effectiveness of our proposed method: actively synthesized classifier-aware pairwise-counterfactual (\textbf{s-CAPC}), we perform counterfactual data augmentation using 10$\%$ counterfactual examples with human-annotated labels as well as 90$\%$  counterfactual examples (a total of 1,000 samples), whose labels are predicted using each method. The error rate on the hold-out counterfactual examples (referred as robustness) as well as on the original test set are shown in Figure \ref{robust1}. 

We can clearly see that (1) the error rate of our proposed method: \textbf{s-CAPC} significantly outperforms other baselines on models' robustness. (2) Comparing PC and CAPC, we can see that CAPC performs slightly better than PC. This indicates that the prediction of the original classifier $f(x), f(c_s(x))$ does provide additional information to help with labels prediction. (3) In addition, we also compare our methods with the extreme case that all the counterfactual examples (100$\%$) are provided human-annotated labels, denoted as (\textbf{human-labels}). Surprisingly, our methods, which only use 10$\%$ human-annotated labels and predict the labels for the other 90$\%$ counterfactual data, achieve comparable performance in improving models' robustness. This sufficiently supports that our proposed methods can effectively predict the labels for counterfactual examples.  
(4) Looking at the error rate on the hold-out original test set, all the methods share a similar performance on SST-2 and our methods are better than other baselines and comparable to human-labels on QQP.

\subsection{How much human-annotated data do we need?}
\begin{figure}[]
\centering
\includegraphics[width=0.9\columnwidth]{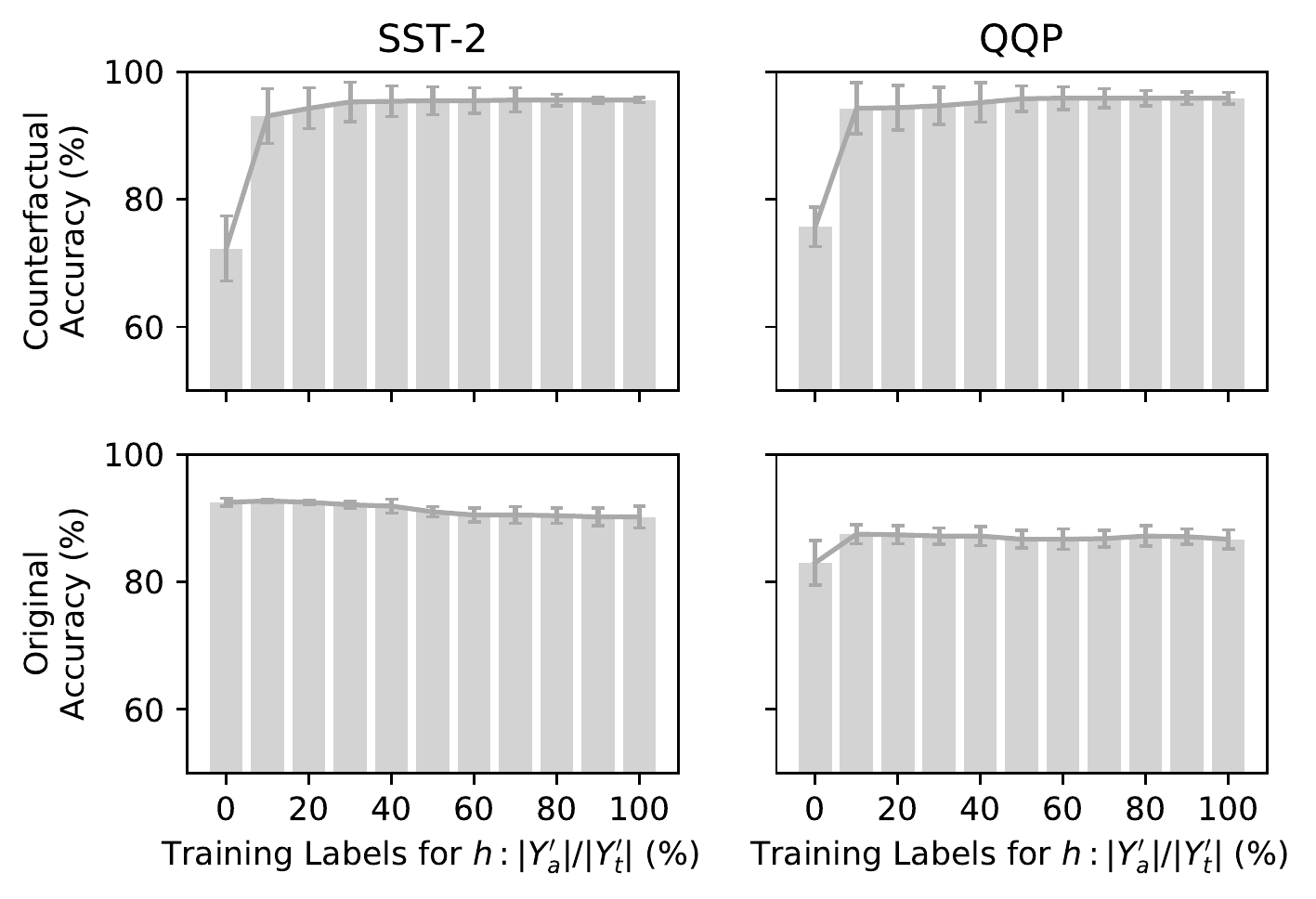}
\caption{\textbf{Impact of training size:} As the number of samples $|Y_a'|$ increases more than $10\%$ in the s-CAPC model , there is not much headroom in counterfactual accuracy, and does not significantly impact the accuracy on the held-out original test dataset on both SST-2 and QQP datasets (overlapping error bounds).}
\label{size}
\vspace{-0.05in}
\end{figure}

To understand the impact of the training data provided to the auxiliary classifier $h$, we increased the \% of data $Y_a'$ provided to the classifier. While this increases costs of annotation, it is important to understand the headroom improvement in counterfactual robustness one would get had they opted for complete human-annotation. Figure \ref{size} shows that across both datasets, the improvement in accuracy and robustness in providing more human annotations to train $h:$ CAPC and subsequently training the model $f:$ RoBERTa-\{SST-2, QQP\} is not significant and hence further demonstrates that, with just 10\% of the augmentation dataset, we can already achieve an improvement comparable to a fully human annotated dataset.
This further confirms our method can achieve high \textit{sample efficiency} in improving models' robustness.

\subsection{Generalization across Counterfactual Types}
We evaluate the generalization of our pairwise counterfactual classifier $h$ by ablating one counterfactual type (e.g negation, quantifier, etc) at a time during training $h$, but still annotate them to generate the augmented training data for $f$. The results are shown in Table ~\ref{type} (rows 2-9).
We see that for the SST-2 task, our approach outperforms existing baselines on counterfactual robustness. This further indicates the importance of learning a counterfactual classifier which captures patterns of label invariance that generalizes across counterfactual templates.
Finally, we evaluate if our generated augmentation dataset can be used to improve \textit{unseen} counterfactual types - ablated while training both $h$ and $f$.
While this is not the goal of our paper, it is useful to understand what types of counterfactuals are captured by our generator and if any overlap between the types of counterfactuals is leveraged.
Table \ref{type} (row 10) shows that our approach is comparable with baselines (rows 2-5 in Table \ref{type}) when a specific counterfactual type is ablated completely from the data augmentation pipeline. This is consistent with existing work \cite{DBLP:journals/corr/abs-2004-15012,huang-etal-2020-counterfactually} and further highlights the need to incorporate diverse types of counterfactuals to perform data augmentation. 

\begin{table*}
\centering
\resizebox{0.9\textwidth}{!}{
\begin{tabular}{l|rrrrrrrr}
\multicolumn{9}{c}{Sliced Error by Counterfactual Type $\%$}\\
\toprule
\textbf{Model}                      & \multicolumn{1}{l}{\textbf{negation}} & \multicolumn{1}{l}{\textbf{quantifier}} & \multicolumn{1}{l}{\textbf{lexical}} & \multicolumn{1}{l}{\textbf{resemantic}} & \multicolumn{1}{l}{\textbf{insert}} & \multicolumn{1}{l}{\textbf{delete}} & \multicolumn{1}{l}{\textbf{restructure}} & \multicolumn{1}{l}{\textbf{shuffle}}\\
\toprule
\textbf{s-CAPC-no-ablation} & 2.20 & 1.81 & 1.94 & 1.40 & 2.01 & 1.75 & 2.01 & 2.12\\
\textbf{no-cda} & 19.12 &	18.10 &	21.40 &	20.65 &	17.54 & 20.99 &	18.32 & 17.42\\
\textbf{invariant}                      & 14.62                                 & 4.82                                    & 4.32                        & 3.10                                    & 7.72                                & 7.83                                & 6.48                                     & 9.24                     \\
\textbf{trust}                                & 12.96                                 & 4.15                                    & 4.73                                 & 3.00                                    & 4.95                                & 12.49                               & 3.74                            & 9.02 \\
\textbf{w-trust}                       & 5.09                                  & 3.55                           & 8.91                                 & 10.60                                   & 7.72                                & 5.57                                & 10.51                                    & 10.60 \\
\textbf{random} & 4.74                         & 4.04                                    & 6.92                                 & 2.22                           & 7.42                                & 5.55                                & 5.72                                     & 4.96  \\
\textbf{training} & 4.53 & 3.53 &	6.32 &	2.62 &	7.24 &	5.32 &	5.83	& 4.83\\
\midrule
\multicolumn{9}{c}{Slice error when counterfactual type is ablated from training $h$}\\
\midrule
\textbf{PC}   & 4.50          & 5.35          & 2.73 & 3.20          & 2.12 & 2.13 & 5.30          & 5.10\\
\textbf{CAPC} & 4.04 & 2.20 & 4.76          & 2.10 & 4.56          & 4.67          & 3.56 & 4.50\\
\textbf{p-CAPC}       & 3.12                                 & 2.01                                   & 2.21                                 & 1.78                                 & 2.66                                & 2.65                                & 2.08                                  & 2.48                               \\
\textbf{s-CAPC}       & \textbf{2.54}                               & \textbf{1.84}                                   & \textbf{2.19}                               & \textbf{1.46}                                  & \textbf{2.07}                                & \textbf{1.86}                              & \textbf{2.02}                                  & \textbf{2.14}                               \\

\midrule
\multicolumn{9}{c}{Sliced error when counterfactual type is ablated from training $h$ and $f$}\\
\midrule
\textbf{s-CAPC}       & 11.17                                 & 13.02                                   & 7.55                                 & 13.33                                   & 4.98                                & 5.76                                & 10.77                                    & 9.01                                 \\

\midrule
  
\end{tabular}
}
\vspace{-0.05in}
\caption{\textbf{Generalization of Counterfactual Types:} Comparison of error rates (\%) sliced by different counterfactual sentence types shows that our approach s-CAPC continues to perform well even when those types are held out during training $h$. However, when we ablate the counterfactual type both while training $f$ and $h$, our approaches performs comparably to the baselines sliced error rates. This shows that $h$ does not just memorize the templates, but training on diverse counterfactual types continues to be important for robustness.}
\label{type}
\end{table*}

\begin{table}[]
    \centering
    \setlength\tabcolsep{2pt}
    \resizebox{0.8\columnwidth}{!}{
    \begin{tabular}{c|cccccc}
    \multicolumn{6}{c}{Test error rate \%}\\
    \toprule
        \textbf{Model} & \textbf{IMDB} & \textbf{Yelp} & \textbf{Amazon} & \textbf{SemEval} & \textbf{IMDB-\scriptsize{cont}} & \textbf{IMDB-\scriptsize{CAD}}\\\toprule
         no-CDA &  9.2 & 15.7 & 20.0 & 15.2 & 7.8 & 13.5\\
         invariant & 11.3 & 15.9 & 21.5 & 15.4 & 8.0 & 13.8\\
         trust & 9.3 & 15.8 & 20.5 & 15.5 & 8.1 & 13.8\\
         w-trust & 9.2 & 15.5 & 20.2 & 15.5 & 8.0 & 13.7\\
         random & 10.4 & 16.3 & 23.8 & 17.2 & 9.5 & 14.3\\
         PC & 8.0 & 14.3 & 18.1 &  14.2 & 7.4 & 12.9\\
         CAPC & \textbf{7.2} & 13.1 & 17.2 & 13.6 & 6.0 & 10.3\\
         p-CAPC & 9.2 & 13.7 & 15.9 & 13.6 & 5.9 & 10.1\\
         s-CAPC & 7.2 & \textbf{10.4} & \textbf{12.9} & \textbf{11.9} & \textbf{5.5} & \textbf{9.9}\\
         \midrule
         domain-trained & 6.7 & 10.0 & 11.7 & 10.8 & 5.4 & 9.5\\\midrule
    \end{tabular}
    }
    \vspace{-0.1in}
    \caption{Out-of-domain reviews: Using data augmentation with SST-2 counterfactuals from the Polyjuice generator and classified using s-CAPC performs comparable to a model trained on within-domain data.}
    \label{tab:ood-reviews}
\end{table}

\subsection{Checklist Evaluation}
\begin{figure}[h!]
\centering
\includegraphics[width=0.8\columnwidth]{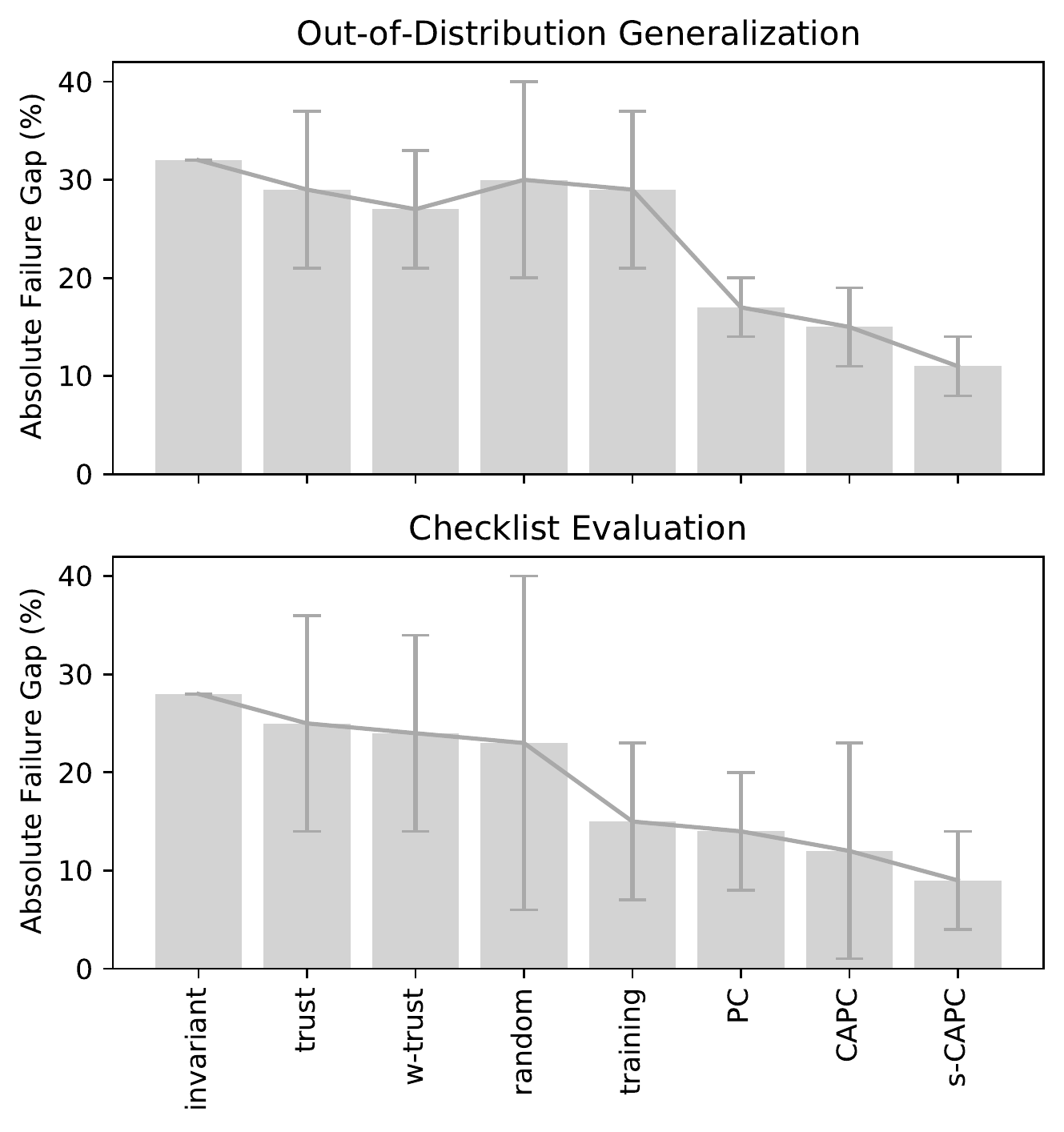}
\vspace{-0.1in}
\caption{ \textbf{Checklist Evaluation - (a) Out of distribution data}: Our methods perform well over different label-invariant distributions with 90\% counterfactual label flips ($y\neq y'$) in the Checklist dataset even when the training distribution has only 10\% counterfactual label flips; \textbf{(b) Model Comparison}:  However, on the original Checklist dataset \cite{ribeiro-etal-2020-beyond}, we achieve a comparable failure gap with the golden error rate to other model-based annotations}
\label{robust2}
\vspace{-0.05in}
\end{figure}

To further validate that the generated labels by our auxiliary model can be used for other tasks, we evaluate it against the labels in CheckList \cite{ribeiro-etal-2020-beyond} which capture other types of counterfactuals. We measure the \emph{Absolute Failure Gap}: $|\epsilon - \epsilon_a|$  computed as the difference between the true error rate $\epsilon$ and the error rate as reported by using our augmented dataset $\epsilon_a$ while evaluating the models and tasks in the CheckList dataset. In Figure \ref{robust2}, we see that even when the training data provided to the auxiliary classifier is synthetically made explicitly label-invariant (90\%), evaluating against counterfactuals with minimal label-invariance (10\%), our model generalizes with a lower failure gap than other augmentation approaches. 
However, on the original Checklist dataset there is no significant improvement in failure gap  compared to reporting the failure gap just on the training data alone.

\subsection{Out-of-Domain Reviews}
To validate that the counterfactuals we augment through our pairwise classifier's annotations have generalizability to 6 out-of-domain datasets, we evaluate the reduction in error rates of the base RoBERTa model when they are trained on the pairwise classifier's data augmentation in Table \ref{tab:ood-reviews}. In the IMDB reviews dataset \cite{maas-etal-2011-learning}, we see an improvement in error rates from 9.2\% without data augmentation to 7.2\% through CAPC and s-CAPC. This out-of-domain error rate is comparable to the error rate obtained by the model trained by \cite{Kaushik2020Learning} after incorporating samples from the counterfactuals drawn from the same distribution as part of the training (6.7\%). In the Yelp reviews too \footnote{https://www.yelp.com/dataset/}, we see a reduction from 15.7\% to 10.4\% whereas other baseline approaches lead to an increase in error rates. In the Amazon reviews, the s-CAPC approach (12.9\%) outperforms the baselines and is comparable to the augmentation from the training split from the Amazon reviews (11.7\%). Similar improvements can be seen on the SemEval \cite{rosenthal-etal-2017-semeval} and IMDB contrast sets (IMDB-cont, IMDB-CAD) \cite{DBLP:journals/corr/abs-2004-02709, Kaushik2020Learning}. Each of these improvements has to be viewed with the context that it was achieved in a more sample efficient manner (1,000 counterfactuals generated from the original SST-2 dataset by Polyjuice) as compared to the in-distribution training approach, where the training data has 3,400 samples from their own respective datasets. This further confirms that training on augmented counterfactuals using a generator and pairwise classifier approach is comparable to human-annotated samples from other domains, while providing us the ability to scale both in terms of domain generalization as well as labeling efficiency.





\section{Limitations}
In this work, we have demonstrated new methods to safely use more diverse counterfactuals and their value, but in taking on this broader goal, we discover a number of further steps that could take the work further forward.
One of the limitations of our paper is that the set of counterfactuals we improved robustness over is limited and restricted to perturbations in the English language. Our analysis indicates the value of using more diverse counterfactual types that require a case-by-case contextual understanding. We show that adding more counterfactual types can be done in a sample efficient manner by using a generator trained to produce counterfactuals. However, this still suffers from the limitation that to extend to more counterfactuals and languages, a classifier which labels them by training on a small set of human annotations is required. Further, we do not investigate the quality of the counterfactuals annotated, and we do not study the performance using more nuanced counterfactuals with low levels of inter-rater agreement. Since we use an auxiliary classifier to label the generated counterfactuals, the risk of label drift remains a clear challenge and we do not control for this label drift based on the certainty of these labels from the auxiliary classifier. Further, a natural drift in concepts based on active exploration might render invalid sentences that are not grammatically or semantically correct, and new methods would be needed to filter based on these text patterns.

As in other generative models, the risk of perpetuating or amplifying biases in the generated text data continues to be important and while we believe counterfactual generation and augmentation can help address such biases, there is also uncertainty in using more flexible, generated counterfactuals. For example, it is quite possible that one of the generated counterfactuals relies on an identify term in the generated sentence, and attributes a negative sentiment spuriously based on prevalent stereotypes in the text corpus. For this reason, we refer the reader to incorporating bias mitigation strategies like \cite{ctf} in addition to improving counterfactual robustness.

While we show generalization across label variance in templates, we cannot guarantee that by learning solely on label invariant counterfactuals, a classifier can generalize over label modifying counterfactuals or obtain the same levels of sample efficiency on harder classification tasks. 
While generators like Polyjuice \cite{polyjuice:acl21} have been evaluated for fluency, diversity, etc., there is a need to evaluate them within the context of a task and its labels.
However, the gains in robustness shown in Figure 3 and Table 2 further illustrate the need for dataset generation in an efficient manner.
As future work, one can also look towards an efficient crowdsourcing strategy that minimizes the gain provided by the pairwise classifiers as each sample in the annotated dataset provides a unique and diverse counterfactual.

\section{Conclusion}
Counterfactual Data Augmentation approaches have been extensively used to train for counterfactual robustness. As the types of counterfactuals - both label-invariant and label-modifying, over which to evaluate natural language models increase, there is a need to adopt a methodology that can scale with increasing types of counterfactuals. We overcome a significant challenge in doing so, by learning an auxiliary pairwise counterfactual classifier that leverages the patterns of counterfactuals produced by vairous generative models. Using only a small amount of human annotated counterfactual samples, we demonstrate that our method can produce a dataset that improves counterfactual robustness comparable to a fully human-annotated dataset.

\bibliography{main}
\bibliographystyle{unsrt}

\end{document}